\pdfoutput=1

\documentclass[11pt]{article}

\usepackage[final]{coling}

\usepackage{times}
\usepackage{latexsym}
\usepackage{amssymb}
\usepackage[T1]{fontenc}

\usepackage[utf8]{inputenc}

\usepackage{microtype}

\usepackage{inconsolata}
\usepackage{xcolor}
\usepackage{booktabs} 
\usepackage{amssymb}
\usepackage{pifont}

\usepackage{tabu}
\usepackage{fontawesome}

\usepackage{url}
\usepackage{soul}
\usepackage{xcolor} 
\usepackage{todonotes}
\definecolor{blue}{rgb}{0,0, 0.6}
\definecolor{dkgreen}{rgb}{0,0.6,0}
\definecolor{gray}{rgb}{0.5,0.5,0.5}
\definecolor{mauve}{rgb}{0.58,0,0.82}
\definecolor{mauve}{rgb}{0,0,0}
\definecolor{black}{rgb}{0,0,0}
\definecolor{SkyBlue}{RGB}{50, 216, 255}
\definecolor{tri}{rgb}{.25,.88,.82}
\definecolor{lilac}{rgb}{0.85,0.64,0.85}
\definecolor{lightblue}{rgb}{0.53, 0.81, 0.98}
\definecolor{lightskyblue}{rgb}{0.53, 0.81, 0.98}
\definecolor{elegantlightrose}{rgb}{0.99,0.79,0.79}

\definecolor{shadowcolor}{rgb}{0.8, 0.8, 0.8}

\newcommand{\sq}{\faCheckSquare}

\title{GenAI Content Detection Task 2: \\
AI vs. Human – Academic Essay Authenticity Challenge}

  \author{Shammur Absar Chowdhury$^1$, Hind Almerekhi$^1$, Mucahid Kutlu$^2$, Kaan Efe Keleş$^3$, \\{\bf  Fatema Ahmad$^1$, Tasnim Mohiuddin$^1$, George Mikros$^4$, Firoj Alam$^1$}\\
  $^1$Qatar Computing Research Institute, HBKU, Qatar, 
  $^2$Qatar University, Qatar \\  
  $^3$TOBB ETU, Türkiye, 
  $^4$Hamad Bin Khalifa University, Qatar \\  
  \texttt{\{shchowdhury, halmerekhi, fakter, mmohiuddin, gmikros, fialam\}@hbku.edu.qa}\\
\texttt{mucahidkutlu@qu.edu.qa}, \texttt{kaanefekeles@etu.edu.tr}\\
\\}

\begin{document}
\maketitle
\begin{abstract}
This paper presents a comprehensive overview of the first edition of the \textit{Academic Essay Authenticity Challenge}, organized as part of the GenAI Content Detection shared tasks collocated with COLING 2025. This challenge focuses on detecting machine-generated {\em vs} human-authored essays for academic purposes. The task is defined as follows: \textit{``Given an essay, identify whether it is generated by a machine or authored by a human.''} The challenge involves two languages: English and Arabic. During the evaluation phase, 25 teams submitted systems for English and 21 teams for Arabic, reflecting substantial interest in the task. Finally, seven teams submitted system description papers. The majority of submissions utilized fine-tuned transformer-based models, with one team employing Large Language Models (LLMs) such as Llama 2 and Llama 3. This paper outlines the task formulation, details the dataset construction process, and explains the evaluation framework. Additionally, we present a summary of the approaches adopted by participating teams. Nearly all submitted systems outperformed the n-gram-based baseline, with the top-performing systems achieving F1 scores exceeding 0.98 for both languages, indicating significant progress in the detection of machine-generated text.

\end{abstract}

\section{Introduction}
\label{sec:introduction}

The rapid progress in Artificial Intelligence (AI) and the proliferation of generative content produced by LLMs have introduced transformative opportunities across various domains — yet they also pose profound challenges \cite{wu2023survey}. One such challenge lies in the detection and prevention of misuse of LLMs in contexts such as fake news, misinformation, disinformation, and academic dishonesty \cite{tang2024science}. For instance, the volume of AI-generated news on misinformation-prone websites surged by 457\% between January 1, 2022, and May 1, 2023, with a corresponding increase of 57.3\% on mainstream platforms \cite{hanley2024machine}. These issues pose substantial barriers to the broader adoption of LLMs, thereby limiting their potential across various applications. Effectively detecting LLM-generated content is crucial for leveraging the capabilities of these models while mitigating associated risks.


Researchers have responded to these challenges through a variety of approaches. Previous methods include classification algorithms designed to distinguish between AI-generated and human-authored text \cite{guo2023close}, as well as watermarking techniques \cite{10.1145/3474085.3475591, He_Xu_Lyu_Wu_Wang_2022, kirchenbauer2023watermark}. These watermarking approaches strategically embed imperceptible signatures within generated texts, enabling model-specific identification while maintaining human-indistinguishable quality. Other recent efforts have focused on the creation of question-answering datasets such as M4 \cite{wang-etal-2024-m4}, generated by humans and ChatGPT in both English and Chinese and the associated shared task \cite{wang2024semeval}.


Within academic settings, concerns surrounding the potential misuse of LLMs have intensified, particularly regarding academic dishonesty involving AI-assisted essay writing and problem-solving. Recent research has made considerable progress in the development of datasets and benchmarking efforts to address these issues. For instance, \citet{yu2023cheat} introduced the CHEAT dataset, which focuses on abstracts from IEEE Xplore, while \citet{wang-etal-2024-m4} developed a comprehensive multilingual dataset. Additionally, \citet{dugan-etal-2024-raid} presented a robust 
dataset designed to address the challenge of detecting machine-generated text.


Despite these efforts, large-scale initiatives in academic contexts remain limited. Hence, this shared task aims to bridge this gap by tackling the task of distinguishing AI-generated essays from human-authored ones. The challenge attracted substantial interest, with 99 teams registered to access the dataset and 56 teams actively participating in the development and evaluation phases. In the evaluation phase, 25 teams submitted systems for English, and 21 teams participated for Arabic. Furthermore, seven teams submitted system description papers. The majority of participating systems employed transformer-based models, while one team utilized state-of-the-art LLMs such as Llama 2 and Llama 3. Notably, most submissions outperformed the traditional n-gram-based baseline, signaling substantial progress in AI-generated content detection methodologies.


The subsequent sections of this paper are structured as follows: Section \ref{sec:related_work} provides a comprehensive review of related work. Section \ref{sec:task_and_dataset} presents the task formulation and dataset setup. Section \ref{sec:results} presents empirical results and offers a comprehensive overview of participating systems. Finally, Section \ref{sec:conclusion} concludes with a summary of findings and future directions.



\section{Related Work}
\label{sec:related_work}


The detection of AI-generated text relies on analyzing statistical patterns and linguistic features that distinguish human and machine writing styles. \citet{Zaitsu2023} highlight that AI-generated text tends to use repetitive sentence patterns and a limited vocabulary, prioritizing clarity over the nuanced variations of human writing. Similarly, \citet{WeberWulff2023} report that such texts often exhibit lower syntactic complexity and reduced lexical diversity, making them identifiable through these markers. Additionally, \citet{Galle2021} report that higher predictability in word \textit{n}-gram is a key indicator of machine generated text. 


Machine learning approaches have become central to AI-generated text detection.  
\citet{Darda2023} explored traditional classification algorithms such as Support Vector Machines (SVM) and Random Forest.  \citet{Vora2023} propose a multimodal approach that uses BERT to analyze syntactic and semantic features of text and CNN architectures for image.
\citet{Mikros2023} investigated using stylometric features and transformer-based models.
Their findings showed that ensemble techniques, particularly those employing majority voting, outperformed individual classifiers. 

There has also been effort to combine different machine learning approaches. For instance, deep learning architectures can extract features from text, while traditional classifiers make predictions based on these features, leveraging the strengths of both techniques \citep{Bhattacharjee2023}. Incorporating user feedback further enhances hybrid models, enabling them to adapt to real-world usage patterns \citep{Rashidi2023}.

Despite advancements in detection methodologies, significant limitations persist. \citet{WeberWulff2023} reveal that many detection tools struggle with high rates of false positives and false negatives, indicating a need for further refinement. According to \citet{Perkins2024}, humans naturally incorporate varying sentence lengths and structures in their writing, creating what researchers call ``burstiness''---a key feature that distinguishes human-authored content from AI-generated text. This variation in writing style, along with occasional grammatical inconsistencies and stylistic irregularities, represents the natural ``imperfections'' that make human writing unique. Interestingly, \citet{Liang2023} found that texts with lower levels of perplexity and coherence---characteristics often found in writing by non-native English speakers---are more likely to be flagged as human-authored. 



Another challenge in AI-generated content detection is  the lack of transparency in models' predictions, reducing their applicability in real-life scenarios, particularly in high-stakes contexts such as academia and forensic applications.  Thus, a number of researchers worked on developing explainable AI (XAI) methods for AI generated text detection. For instance, \citet{Shah2023} develop an XAI model using stylistic features. \citet{Wu2023} proposes a hybrid approach that combine statistical analysis with machine learning techniques. Additionally, the integration of user feedback into hybrid models may facilitate the development of more adaptive systems that can learn from usage patterns \citep{Rashidi2023}.

\begin{table*}[h!]
\centering
\begin{tabular}{p{0.15\textwidth}p{0.8\textwidth}}
\toprule
\textbf{{\small System Prompt}} &
\texttt{You are a \textbf{\{study\_level\}} student from \textbf{\{country\}}, preparing for the TOEFL exam. Your English proficiency level is \textbf{\{proficiency\_level\}}. Your task is to write a well-structured TOEFL essay in response to the given prompt. Ensure your essay is clear and coherent, following the standard essay format: an introduction, body paragraphs, and a conclusion. Focus on presenting your ideas logically, using appropriate language, and providing relevant examples to support your arguments. Aim to demonstrate your proficiency in English through organized thought and effective communication.
} \\ \midrule
\textbf{{\small User Prompt}} &

\texttt{Do you agree or disagree with the following statement:}
\textbf{"\{statement\}"}

\texttt{Write a well-structured essay expressing your opinion. Be sure to use specific reasons and examples to support your viewpoint.}

\texttt{The essay should be between \textbf{\{min\_length\}} and \textbf{\{max\_length\}} words in length.}

\texttt{Please provide only an essay and in a JSON object. No additional text or explanation.}

\textbf{\{"essay": "your essay"\}}
\\ \bottomrule
\end{tabular}
\caption{Example of \textit{System} and \textit{User Prompts} for training and validation in English essay generation. Similar prompts were used for Arabic essays. Variables include study\_level =\{`pre-university',`university'\}, proficiency\_levels=\{`low',`medium','high'\}, country\_list=\{`Arabic', `German', `French', `Hindi', `Italian', `Japanese', `Korean', `Spanish', `Telugu', `Turkish', `Chinese'\}. For Arabic prompts, an additional variable, nativity=\{`native',`non-native'\} is used.  }
\label{tab:train_prompts}
\end{table*}

\section{Task and Dataset}
\label{sec:task_and_dataset}

\subsection{Task Definition} 
The main objective of the task is to detect whether the given candidate essay is AI-generated or human-written. Given the input essay $\mathbf{e}$, the task is to design a text detector $\mathcal{D}(\mathbf{e})$, such that the model outputs label indicating AI-generated or Human-authored content. For this edition, we designed the task as binary classification problem. 



\subsection{Datasets}
The task aims to develop a system specifically designed for detecting AI generated text in academic essays. The dataset comprises essays authored by both native and non-native speakers, alongside AI-generated content. A significant challenge in this task was collecting authentic human-authored academic essays while addressing the following considerations:
\begin{itemize}
\item  Ensuring author privacy, obtaining informed consent, and ethically sourcing the content.  
\item Verifying that the collected essays were genuinely authored by humans, free from any AI interference or plagiarism.  
\item  Acquiring a diverse set of essays representing different academic levels and cultural backgrounds to ensure inclusivity in the dataset.  
\end{itemize}
For the task, we focused on two languages: English and Arabic. For each language, we provided training, validation, dev-test, and the final test sets, which included human-authored and AI-generated texts. We released these data splits in two phases -- \textit{(i) Development phase} -- we released the training, validation, and mock test data (dev-test); \textit{(ii) Evaluation phase} -- we released the final test set which is used to rank the submitted system. Below, we discuss the dataset design for the development and final evaluation phases, respectively.

\subsection{Development Phase} 
During the development phase we have released training, validation, and dev-test. For this phase, we first collected human-authored essays and essay topics. To create the data splits, we carefully designed each set to ensure unique essay topics, avoiding overlap between training, validation, and dev-test datasets.

Furthermore, within each split, we manually categorized the essay topics based on their thematic similarity. This classification is used to assign topics for generating essays using LLMs, and the rest is reserved exclusively for selecting human-authored essays from various existing datasets mentioned below. The final statistics of the dataset released in this phases are presented in Table \ref{tab:dp_datastats}.



\paragraph{Human-authored Essay} 
The human-authored data was sourced from different language assessment datasets, including examinations like IELTS, and TOEFL among others. To ensure the authenticity of human-authored content, we selected essays that were either handwritten or composed in a supervised classroom setting, explicitly to make sure that none of the texts were created with the assistance of generative technologies or online articles. This approach was designed to maintain the integrity of the datasets and accurately represent human academic writing. 

\noindent For the English, we collected essay statements (essay prompt) and essays from:
\begin{itemize} 
    \item \textbf{IELTS Writing Scored Essays} Dataset\footnote{\url{https://www.kaggle.com/datasets/mazlumi/ielts-writing-scored-essays-dataset}} contains 1200 academic essays for varieties of prompts. Each essays are accompanied by the examiners' feedback along with scores
    \item \textbf{ETS Corpus of Non-Native Written English} corpus\footnote{\url{https://catalog.ldc.upenn.edu/LDC2014T06}} contains 12,100 academic essays, written addressing eight different prompts, by non-native speakers from 11 different countries, as part TOEFL English proficiency exam. The dataset includes the speaker's native language along with scores they obtained for the corresponding essays. While the dataset was originally designed for native language identification tasks, its rich collection of academic essays, makes it highly suitable for supporting our AI-generated text detection efforts. 
\end{itemize}

\noindent As for the Arabic subtask, the datasets we use are the following:
\begin{itemize} 
    \item \textbf{Arabic Learner Corpus (ALC)}\footnote{\url{https://www.arabiclearnercorpus.com}} \cite{wrro75470} includes 1,197 essays written by both native and non-native Arabic pre-university/university speakers from 67 nationalities. The dataset includes speakers' nationality along with the information if the essay was written in class or as homework. For the task, we only selected in-class essays, manually excluded off-topic essays, and reviewed the essays for any corrections. 
    
    \item \textbf{Qatari Corpus of Argumentative Writing (QCAW) dataset}\footnote{\url{https://catalog.ldc.upenn.edu/LDC2022T04}} \cite{zaghouani-etal-2024-qcaw} is a collection of 195 argumentative essays written by native Arabic undergraduate students. The prompts given to the student were inspired by TOEFL writing exercises~\cite{AhmedZhangRezkZaghouani+2023+183+215}. 
    \item \textbf{The CERCLL corpus}\footnote{\url{https://cercll.arizona.edu/arabic-corpus/}} includes $\approx$ 270 essays written by non-native (L2) and heritage Arabic speakers.\footnote{The original dataset is available in pdf format.} The dataset includes information about the speakers' proficiency, along with the type -- L2 {\em vs} heritage speakers. The dataset covers a wide range of topics and multiple genres, including description, narration, and instruction essays.   
\end{itemize}


\begin{table*}[!ht]
\begin{tabular}{p{15.5cm}} \toprule
\small
\texttt{You are tasked with generating creative and rigorous academic essays.}

\texttt{Here’s how:}

\texttt{1) Topics Selection: You are provided with a set of topics: <<<\textbf{20 random topics}>>>. First, choose one topic at random from this list.}

\texttt{2) Generate Related Topics: Based on the chosen topic, create 10 new topic ideas. These should be different from the chosen topic but related in a way that someone interested in the initial topic might also find these new ideas engaging.}

\texttt{3) Select Final Topic: From the 10 new topics, pick one at random to focus on.}

\texttt{4) Choose a Profession: List 10 random professions that are entirely unrelated to the final topic, ensuring that they come from different fields or disciplines. These professions should be distinct enough that their practitioners would not typically engage with or have knowledge about the topic. Then, select one profession at random from this list.}

\texttt{5) Choose a Writing Style: List 10 distinct writing styles (e.g., persuasive, narrative, descriptive) and choose one at random.}

\texttt{6) Essay Writing: Write an academic and creative essay on the chosen topic. This essay should be written from the perspective of someone in the chosen profession and in the selected writing style. Do not ever mention the chosen profession or writing style in the essay itself. Do not include any personal opinions or experiences with regarding to the profession in the essay. Do not mention anything about the chosen profession whatsoever.}

\texttt{Your output should be in JSON format, structured as follows:}

\texttt{\{
    "selected\_topic": "<randomly selected topic from the given topics>",
    "generated\_topics": [
        "<generated topic 1>",
        "<generated topic 2>",
        "...",
        "<generated topic 10>"
    ],
    "final\_topic": "<randomly selected topic from generated\_topics>",
    "professions": [
        "<profession 1>",
        "<profession 2>",
        "...",
        "<profession 10>"
    ],
    "selected\_profession": "<randomly selected profession from professions>",
    "writing\_styles": [
        "<style 1>",
        "<style 2>",
        "...",
        "<style 10>"
    ],
    "selected\_writing\_style": "<randomly selected style from writing\_styles>",
    "essay": "<generated essay>"
\}}

\texttt{Please proceed with this format to generate a fully structured JSON output. Remember to keep the content diverse and creative throughout the process. The essay should be comprehensive, detailed, and reflective of rigorous academic standards. The essay must be multiple paragraphs long (at least 1 page's worth). Return only the valid JSON output and nothing else. Good luck!} \\ \bottomrule
\end{tabular}
\caption{\textbf{Freehand prompt} used to generate AI generated essays for the final test set.}
\label{tab_prompts_free}
\end{table*}

\begin{table*}
\begin{tabular}{p{15.5cm}} \toprule
\small
\texttt{Thoroughly rewrite the provided academic essay to enhance clarity, diversity in sentence structure, and vocabulary richness, all while maintaining the original meaning and intent. Your goal is to produce a refined and nuanced version of the text.}

\texttt{Aim to increase the essay's length by adding substantial elaborations, exploring various perspectives, and providing comprehensive explanations that will offer a deeply layered and extensive output.}

\texttt{Deliver the output exclusively in JSON format with a single key "text" as shown below, ensuring that no additional information or comments are included:}

\texttt{\{\{
    "text": "<rewritten\_and\_greatly\_expanded\_academic\_essay>"
\}\}}

\texttt{Here is the passage to rewrite and extensively expand:}

\texttt{<<<original\_passage\_start>>>}
\texttt{\{\textbf{the passage to be paraphrased}\}}
\texttt{<<<original\_passage\_end>>>} \\ \bottomrule
\end{tabular}
\caption{\textbf{Paraphrasing prompt} used to generate AI generated essays for the final test set.
}
\label{tab_prompts1}
\end{table*}

\paragraph{AI-generated Essay} 
The generated essays, for both languages, utilized seven state-of-the art LLMs including: GPT-3.5-Turbo (2023-03-15-preview), GPT-4o (2024-08-06), GPT-4o-mini (2024-07-18)~\cite{openai2024gpt4omini}, Gemini-1.5~\cite{gemini2024}, phi3.5,\footnote{\url{https://huggingface.co/microsoft/Phi-3.5-mini-instruct}} Llama-3.1 (8B)~\cite{abdin2024phi}, and Claude-3.5.\footnote{\url{https://www.anthropic.com/news/claude-3-5-sonnet}} 
To produce these essays, we designed the prompts by utilizing a selected subset of essay statements from the aforementioned datasets. The designed prompts included detailed instructions to emulate human writing styles, specify essay length requirements, and incorporate predefined personas reflecting various factors such as nativity and/or language proficiency, following the metadata and statistics obtained from the human-authored essay collections. This approach ensured the generation of essays that closely resemble real-world human writing in both style and content. An example of such a prompt is shown in Table \ref{tab:train_prompts}.

\begin{table*}[ht]
\centering
\small
\begin{tabular}{l|p{12cm}}
\toprule
\textbf{Question Type} & \textbf{Example Statements} \\ \midrule
\textit{Agree or Disagree} & Do you agree or disagree with the following statement? People should be encouraged to take risks, even if there is a chance of failure. Use specific reasons and examples to support your answer. \\ \hline
\textit{Preference} & Some people prefer to spend their money on experiences, such as travel or concerts, while others prefer to save for physical possessions, such as a car or a home. Which approach do you prefer, and why? Use specific reasons and examples to support your choice. \\ \hline
\textit{If/Imaginary Situations} & If you could have any superpower, such as the ability to fly or become invisible, which one would you choose, and why? Use specific reasons and examples to explain your answer. \\ \hline
\textit{Advan. and Disadvan.} & What are the advantages and disadvantages of living in a large city? Use specific reasons and examples to support your answer. \\ \hline
\textit{Descriptive} & Describe a memorable trip you have taken and explain what made it special. Use specific details to support your response. \\ \bottomrule
\end{tabular}
\caption{Examples of different question types and corresponding essay statements (prompts).}
\label{tab:question_types}
\end{table*}

\begin{table}[h]
\centering
\scalebox{0.9}{
\begin{tabular}{l|rrrr}
\toprule
\textbf{Label} & \textbf{Train} & \textbf{Valid} & \textbf{Dev-Test} & \textbf{Total} \\ \midrule
\multicolumn{5}{c}{\textbf{English}} \\ \midrule
\textit{AI} & 925 & 299 & 712 & 1,936 \\
\textit{Human} & 1,145 & 182 & 174 & 1,501 \\
\textbf{Total} & 2,070 & 481 & 886 & 3,437 \\
\midrule
\multicolumn{5}{c}{\textbf{Arabic}} \\ \midrule
\textit{AI} & 1,467 & 391 & 369 & 2,227 \\
\textit{Human} & 629 & 1,235 & 500 & 2,364 \\
\textbf{Total} & 2096 & 1,626 & 869 & 4,591 \\ \bottomrule
\end{tabular}}
\caption{Development phase: dataset and label distribution }
\label{tab:dp_datastats}
\end{table}

\subsection{Evaluation Phase} 
For the evaluation, we designed and developed a novel dataset, the \textbf{G}enerated and \textbf{R}eal \textbf{A}cademic \textbf{C}orpus for \textbf{E}valuation (GRACE), which includes both human-authored and AI-generated essays in English and Arabic.
\subsubsection{Data Collection} For designing the human-authored portion of the dataset, we began by carefully designing test set essay statements aligned with those used in development phase topics. 
We selected five different essay types, and under each type, we created several essay statements (see Table \ref{tab:question_types} for examples). The topics include social influence \& technology, lifestyle choices \& preferences, cultural \& global perspective, environmental \& societal responsibility, and personal growth \& experience.


\paragraph{Essay Writing by Recruited Participants:}
We then recruited\footnote{We use a third-party company for the reward money. The amount was decided based on the standard local rate for data annotation.} university students, both monolingual and bilingual, contribute to the essay writing. The participants were provided with a list of essay statements in their respective languages (either English or Arabic) and were asked to complete each essay within 30 minutes. They were instructed to limit the essays to 350–500 words and ensure they included an introduction, main arguments, and a conclusion. The essays must be written in Modern Standard Arabic (MSA) for Arabic, or in formal English for the English essays.  

\paragraph{Collected Essay Assignments:}
Additionally, we collected previously submitted English \textit{essay assignments} from university students to enrich the dataset.

\noindent\textit{Anonymization of Personal Information} In the collected \textit{essay assignments} we noticed that there were some information containing mentions of entities.  
Therefore, we anonymized them to ensure the removal of any information that could directly or indirectly identify the author or reveal any private information about an entity that is not publicly known. This process was essential to uphold privacy standards and ethical considerations.

\noindent To achieve this, we followed these guidelines:
\begin{itemize} 
    \item \textit{Author Identification Removal}: Any mention of names, addresses, affiliations, or specific details that could identify the essay's author was redacted.
    \item \textit{Private Entity Information}: Any references to non-public entities, such as organizations, businesses, or private individuals mentioned in the essays, were removed or replaced with generic terms.
    \item \textit{Sensitive Content}: Sensitive information, such as health conditions, financial details, or other personal data, was also removed to ensure privacy.
    \item \textit{Consistency}: Replacement terms were standardized (e.g., ``[NAME]``, ``[ADDRESS]'', ``[ORGANIZATION]'') to maintain consistency throughout the dataset.
\end{itemize}

\noindent A team of five trained annotators was recruited to carry out this task. Each annotator was provided with clear anonymization guidelines and examples to ensure consistency and accuracy. Such anonymization steps ensure that the dataset meets ethical standards for research.

\subsubsection{Data Generation}
For the AI-generated essays, we followed two distinct methodologies: 
\begin{itemize}
    \item \textit{Freehand Generation:} An instruct-tuned LLM, namely gpt-4o, independently generated essays using the \textit{Freehand Generation Prompt} shown in Table  \ref{tab_prompts_free}. The prompt was designed to ensure diverse outputs. We were inspired by the prompting techniques proposed by \citet{chen2024genqageneratingmillionsinstructions}.
    \item \textit{Paraphrasing Human-Written Text:} Using the \textit{Paraphrasing Prompt} shown in Table \ref{tab_prompts1}, human-authored essays were rephrased by an instruct-tuned LLM, namely claude-3.5
    to generate stylistically varied yet semantically equivalent AI-written versions. The resulting text comprises a mix of human-written and AI-generated content, designed to challenge the effectiveness of detection methods.
\end{itemize}

\begin{table}[ht]
\centering
\setlength{\tabcolsep}{6pt} 
\scalebox{1.}{%
\begin{tabular}{l|rrr}
\toprule
\textbf{Category} & \textbf{English} & \textbf{Arabic} & \textbf{Total} \\ \midrule
AI (Free) & 400 & 100 & 500 \\ 
AI (Para) & 365 & 98 & 463 \\ \hline
Human & 365 & 95 & 460 \\ \hline
\textbf{Total} & 1,130 & 293 & 1,423 \\ \bottomrule
\end{tabular}
}
\caption{Distribution of essays by \textit{category} and \textit{language} across the test set. Free - freehand generation, Para - paraphrasing-based generation. }
\label{tab:category_language_distribution}
\end{table}





The final GRACE dataset comprises a balanced distribution of human-written and AI-generated essays.  Table \ref{tab:category_language_distribution} provides a detailed breakdown across languages and generation methods.

\subsection{Baseline and Evaluation Setup}
\label{ssec:evaluation_setup}

\subsubsection{Baseline}
For all languages, we train an n-gram (unigram, $n=1$) based baseline model. We transformed the texual content of the essays into a TF-IDF (Term Frequency-Inverse Document Frequency) representation with a maximum of 10k features. A Support Vector Machine (SVM) classifier is then trained on this feature representation to evaluate its performance.

\subsubsection{Evaluation Setup}
The task was organized into two phases, corresponding to the previously described dataset development process:
\begin{itemize} 
    \item \textbf{Development phase}: We released the train and validation subsets, and participants submitted runs on the dev-test set through a competition on Codalab.\footnote{\href{https://codalab.lisn.upsaclay.fr/competitions/20118}
    {https://codalab.lisn.upsaclay.fr/competitions/20118}}    
    \item \textbf{Evaluation phase}: We released the official test subset -- GRACE, and the participants were given four days 
    to submit their final predictions through the same Codalab competition URL. Only the latest submission from each team was considered official and was used for the final team ranking.
\end{itemize}

\subsubsection{Evaluation Measure:}
We measure the performance of the participating systems using accuracy, macro- precision, recall and F1 measure. However, official ranking was based on macro-F1.

\begin{table*}[!ht]
\centering
\setlength{\tabcolsep}{3pt} 
\scalebox{0.85}{%
\begin{tabular}{@{}lrrrrr|lrrrrr@{}}
\toprule
\multicolumn{6}{c|}{\textbf{Arabic}} & \multicolumn{6}{c}{\textbf{English}} \\ \midrule
\multicolumn{1}{c}{\textbf{Team}} & \multicolumn{1}{c}{\textbf{Acc}} & \multicolumn{1}{c}{\textbf{P}} & \multicolumn{1}{c}{\textbf{R}} & \multicolumn{1}{c}{\textbf{F1}} & \multicolumn{1}{c|}{\textbf{Rank}} & \multicolumn{1}{c}{\textbf{Team}} & \multicolumn{1}{c}{\textbf{Acc}} & \multicolumn{1}{c}{\textbf{P}} & \multicolumn{1}{c}{\textbf{R}} & \multicolumn{1}{c}{\textbf{F1}} & \multicolumn{1}{c}{\textbf{Rank}} \\ \midrule
IntegrityAI & 0.986 & 0.990 & 0.979 & 0.984 & 1 & CMI-AIGCX & 0.999 & 0.999 & 0.999 & 0.999 & 1 \\
USTC-BUPT & 0.976 & 0.983 & 0.963 & 0.972 & 2 & starlight & 0.997 & 0.998 & 0.996 & 0.997 & 2 \\
starlight & 0.969 & 0.964 & 0.966 & 0.965 & 3 & saehyunMa & 0.994 & 0.995 & 0.990 & 0.993 & 3 \\
CMI-AIGCX & 0.969 & 0.966 & 0.964 & 0.965 & 4 & Fsf & 0.994 & 0.995 & 0.990 & 0.993 & 4 \\
apricity & 0.966 & 0.969 & 0.953 & 0.960 & 5 & 1-800 & 0.991 & 0.987 & 0.993 & 0.990 & 5 \\
RA & 0.962 & 0.956 & 0.959 & 0.957 & 6 & Tesla & 0.988 & 0.983 & 0.989 & 0.986 & 6 \\
1-800 & 0.959 & 0.961 & 0.945 & 0.952 & 7 & apricity & 0.988 & 0.983 & 0.989 & 0.986 & 7 \\
Lkminnow & 0.956 & 0.943 & 0.959 & 0.950 & 8 & small & 0.984 & 0.981 & 0.983 & 0.982 & 8 \\
alpaca0000001 & 0.949 & 0.937 & 0.948 & 0.942 & 9 & jojoc & 0.982 & 0.975 & 0.985 & 0.980 & 9 \\
jojoc & 0.949 & 0.939 & 0.946 & 0.942 & 10 & EssayDetect & 0.978 & 0.968 & 0.984 & 0.975 & 10 \\
small & 0.945 & 0.938 & 0.938 & 0.938 & 11 & ShixuanMa & 0.976 & 0.968 & 0.979 & 0.973 & 11 \\
jebish7 & 0.945 & 0.945 & 0.929 & 0.937 & 12 & RA & 0.973 & 0.975 & 0.964 & 0.969 & 12 \\
EssayDetect & 0.942 & 0.949 & 0.919 & 0.932 & 13 & alpaca0000001 & 0.956 & 0.940 & 0.967 & 0.951 & 13 \\
nits\_teja\_srikar & 0.922 & 0.943 & 0.882 & 0.904 & 14 & Lkminnow & 0.932 & 0.913 & 0.943 & 0.925 & 14 \\
Mashixuan & 0.898 & 0.877 & 0.911 & 0.889 & 15 & IntegrityAI & 0.880 & 0.864 & 0.911 & 0.873 & 15 \\
Sinai & 0.829 & 0.821 & 0.866 & 0.822 & 16 & USTC-BUPT & 0.878 & 0.922 & 0.812 & 0.842 & 16 \\
Vasudha & 0.816 & 0.796 & 0.831 & 0.804 & 17 & jebish7 & 0.847 & 0.908 & 0.763 & 0.794 & 17 \\
ShixuanMa & 0.758 & 0.783 & 0.818 & 0.754 & 18 & CNLP-NITS-PP & 0.777 & 0.784 & 0.825 & 0.771 & 18 \\
gaoyf & 0.608 & 0.720 & 0.707 & 0.607 & 19 & Mashixuan & 0.742 & 0.778 & 0.809 & 0.739 & 19 \\
CNLP-NITS-PP & 0.590 & 0.557 & 0.563 & 0.557 & 20 & nits\_teja\_srikar & 0.773 & 0.875 & 0.649 & 0.658 & 20 \\
halcyonized & 0.495 & 0.488 & 0.487 & 0.475 & 21 & Vasudha & 0.517 & 0.700 & 0.643 & 0.509 & 21 \\
\textit{Baseline} & 0.474 & 0.480 & 0.477 & 0.461 & - & Mahavir\_IIITA & 0.512 & 0.683 & 0.634 & 0.504 & 22 \\
 &  &  &  &  &  & \textit{Baseline} & 0.495 & 0.494 & 0.494 & 0.478 & - \\
 &  &  &  &  &  & halcyonized & 0.493 & 0.494 & 0.493 & 0.477 & 23 \\
 &  &  &  &  &  & gaoyf & 0.391 & 0.523 & 0.514 & 0.374 & 24 \\
 &  &  &  &  &   & Sinai & 0.354 & 0.602 & 0.519 & 0.298 & 25 \\ \bottomrule
\end{tabular}
}
\caption{The official results for Arabic and English are ranked based on the official metric: macro-F1. Teams that submitted a system description paper are indicated in bold.}
\label{tab:results}
\end{table*}

\section{Results and Overview of the Systems}
\label{sec:results}

In Table \ref{tab:results}, we present the results of participants' systems for both Arabic and English including baseline.
For Arabic, all systems outperformed the n-gram baseline, whereas, for English, three teams performed below the baseline. The task generated significant interest, with 56 teams registering to participate. However, the number of system submissions was nearly halved, and ultimately, only five teams submitted system description papers.
In Table \ref{tab:team_comparison}, we provide an overview of the participating systems for which a description paper was submitted. For Arabic top team, \textbf{IntegrityAI}~\cite{genai-detect:2025:task2:IntegrityAI}, fine-tuned Electra model. For English top team, \textbf{CMI-AIGCX} \cite{genai-detect:2025:task:CMI-AIGCX}, used LLMs (Llama 2 and 3) and also fine-tuned XLM-roberta model.  


\begin{table}[h]
\centering
\setlength{\tabcolsep}{2pt} 
\scalebox{0.8}{%
\begin{tabular}{l|cc|cccccccccc|cc}
\toprule
\multicolumn{1}{c}{\textbf{Team}} & \multicolumn{2}{|c}{\textbf{Lang.}} & \multicolumn{10}{|c}{\textbf{Models}} & \multicolumn{2}{|c}{\textbf{Misc}} \\ \midrule
\rotatebox{90}{\textbf{}} & \rotatebox{90}{Arabic} & \rotatebox{90}{English} & \rotatebox{90}{LLama2} & \rotatebox{90}{LLama3} & \rotatebox{90}{BERT} & \rotatebox{90}{RoBERTa} & \rotatebox{90}{XLM-r} & \rotatebox{90}{ALBERT} & \rotatebox{90}{DistilBERT} & \rotatebox{90}{DeBERTa} & \rotatebox{90}{Electra} & \rotatebox{90}{AraBERT} & \rotatebox{90}{Prep.} & \rotatebox{90}{Info.} \\ 
\midrule
IntegrityAI   & 1 & 15 &        &         &         &         &         &         &            &         & \sq &         & \sq & \sq \\ 
CMI-AIGCX     & 4 & 1 & \sq & \sq &         &         & \sq &         &            &         &         &         & \sq &          \\ 
Tesla         &           & 6 &        &         &         &         &         &         &            &         &         &         &          &          \\ 
EssayDetect   & 13 & 10 &        &         & \sq & \sq & \sq & \sq & \sq  &         &         &         &          &          \\ 
RA            & 6 & 12 &        &         &         & \sq &         &         &            & \sq &         & \sq &          &          \\ 
\bottomrule
\end{tabular}
}
\caption{Overview of the approaches. The numbers in the language box refer to the position of the team in the official ranking. Prep.: Preprocessing. Info.: Info. Extraction.}
\label{tab:team_comparison}
\end{table}

Team~\textbf{IntegrityAI}~\cite{genai-detect:2025:task2:IntegrityAI} fine-tuned ELECTRA-small for English and AraELECTRA-base for Arabic to balance high performance with computational efficiency. Stylometric features, including word count, sentence length, and vocabulary richness, were incorporated to enhance detection capabilities. The lightweight models achieved F1-scores of 0.985 for English and 0.984 for Arabic, demonstrating the effectiveness of combining transformer-based architectures with stylometric analysis. The system was further optimized for deployment on GPUs with moderate memory capacity, ensuring both efficiency and accessibility. Larger models, such as ELECTRA-large, were also tested, achieving an F1-score of 0.997 for English, demonstrating the potential for even greater accuracy with additional computational resources.


Team \textbf{CMI-AIGCX}~\cite{genai-detect:2025:task:CMI-AIGCX} proposed a method leveraging the Llama-3.1-8B model as a proxy to capture the semantic feature of each token in the text. These token representations were subsequently used to train a model. 
Instead of fine-tuning an LLM, they leveraged multilingual knowledge and trained a model to enhance detection performance. Their approach demonstrated that using a proxy model with diverse multilingual knowledge can effectively detect machine-generated text across multiple languages, regardless of model size. For English, an F1 score of 0.999 was achieved, securing first place out of 25 teams. For Arabic, an F1 score of 0.965 was obtained, which ranked fourth among 21 teams.

Team \textbf{Tesla}~\cite{genai-detect:2025:task2:Tesla} extracted a comprehensive set of features encompassing style, language complexity, bias, subjectivity, and emotion. These features were used to train four machine learning algorithms: Logistic Regression, Random Forest, Randomized Decision Trees (Extra Trees), and XGBoost, leveraging diverse approaches to optimize detection performance. Their methods ranked 6th on the leaderboard for the English subtask, achieving an 
F1-score of 0.986.

Team \textbf{EssayDetect}~\cite{genai-detect:2025:task:EssayDetect} proposed a fusion model by integrating pre-trained language model embeddings with stylometric and linguistic features to improve classification accuracy. The contributions were threefold: \textit{(i)} LIME was utilized to identify and highlight highly discriminative features, \textit{(ii)} focal loss was employed to address class imbalance, and \textit{(iii)} layer-wise freezing was implemented during fine-tuning to preserve core linguistic representations in the lower layers while enabling the higher layers to capture task-specific stylistic differences in essays.

Team \textbf{RA}~\cite{genai-detect:2025:task:RA} fine-tuned several models for English, including RoBERTa, XLM-RoBERTa, mBERT, and DeBERTa. Similar performance was observed across all models on the validation set, except for mBERT, which exhibited slightly lower performance. For Arabic, AraBERT, ArBERT, and MarBERT were fine-tuned on the full dataset. AraBERT consistently demonstrated superior performance in terms of F1-score across both languages. The models consistently exceeded both the mean and median scores across tasks, achieving an F1-score of 0.969 in classifying AI-generated essays in English and 0.957 in Arabic.

\section{Conclusion and Future Work}
\label{sec:conclusion}
We presented an overview of the shared task on the \textit{Academic Essay Challenge}. The task attracted significant attention, with a total of 56 teams registering to participate in the development and evaluation phases. Of these, 21 teams submitted official results on the test set for Arabic, and 25 teams did so for English. Finally, seven teams submitted task description papers. Most systems fine-tuned transformer-based language models; however, several teams also incorporated additional features, such as style, language complexity, bias, subjectivity, and emotion. For both languages, the top-performing teams achieved F1 scores above 0.98.

\section*{Limitations}
A major limitation of the dataset is its small size, particularly for Arabic, which restricts the development of more robust models. The challenging nature of academic essay collection is reflected in the limited dataset size. Future studies could focus on curating larger datasets to enable the creation of more challenging tasks and the development of more robust models. 

\section*{Ethical Considerations}
The datasets used in the shared task may reflect subjective biases or perspectives of the essay authors, even though they followed the provided instructions. Importantly, the datasets do not include any personal information, and no such information was collected during the data curation process. Therefore, we do not anticipate any ethical concerns related to privacy. Furthermore, the dataset was shared only with participants who signed an agreement, ensuring responsible use of the dataset. 

\section*{Acknowledgments}
The work of F. Alam and G. Mikrosis partially supported by HBKU signature grant (HBKU-OVPR-SRG-02-2). The findings achieved herein are solely the responsibility of the authors.

\bibliography{bib/bibliography}

\begin{thebibliography}{35}
\providecommand{\natexlab}[1]{#1}

\bibitem[{Abdin et~al.(2024)Abdin, Jacobs, Awan, Aneja, Awadallah, Awadalla, Bach, Bahree, Bakhtiari, Behl et~al.}]{abdin2024phi}
Marah Abdin, Sam~Ade Jacobs, Ammar~Ahmad Awan, Jyoti Aneja, Ahmed Awadallah, Hany Awadalla, Nguyen Bach, Amit Bahree, Arash Bakhtiari, Harkirat Behl, et~al. 2024.
\newblock Phi-3 technical report: A highly capable language model locally on your phone.
\newblock \emph{arXiv preprint arXiv:2404.14219}.

\bibitem[{Agrahari et~al.(2025)Agrahari, Jayant, Kumar, and Sanasam}]{genai-detect:2025:task:EssayDetect}
Shifali Agrahari, Subhashi Jayant, Saurabh Kumar, and Ranbir Sanasam. 2025.
\newblock {Team EssayDetect at GenAI Detection Task 2: Guardians of Academic Integrity: Multilingual Detection of AI-Generated Essays}.
\newblock In \emph{Proceedings of the 1st Workshop on GenAI Content Detection (GenAIDetect)}, Abu Dhabi, UAE. International Conference on Computational Linguistics.

\bibitem[{Ahmed et~al.(2023)Ahmed, Zhang, Rezk, and Zaghouani}]{AhmedZhangRezkZaghouani+2023+183+215}
Abdelhamid~M. Ahmed, Xiao Zhang, Lameya~M. Rezk, and Wajdi Zaghouani. 2023.
\newblock \href {https://doi.org/doi:10.1515/csh-2023-0012} {Building an annotated l1 arabic/l2 english bilingual writer corpus: The qatari corpus of argumentative writing (qcaw)}.
\newblock \emph{Corpus-based Studies across Humanities}, 1(1):183--215.

\bibitem[{AL-Smadi(2025)}]{genai-detect:2025:task2:IntegrityAI}
Mohammad AL-Smadi. 2025.
\newblock {IntegrityAI at GenAI Detection Task 2: Detecting Machine-Generated Academic Essays in English and Arabic Using ELECTRA and Stylometry}.
\newblock In \emph{Proceedings of the 1st Workshop on GenAI Content Detection (GenAIDetect)}, Abu Dhabi, UAE. International Conference on Computational Linguistics.

\bibitem[{Alfaifi and Atwell(2013)}]{wrro75470}
AYG Alfaifi and ES~Atwell. 2013.
\newblock \href {https://eprints.whiterose.ac.uk/75470/} {Arabic learner corpus v1: A new resource for arabic language research}.
\newblock In \emph{Second Workshop on Arabic Corpus Linguistics}.

\bibitem[{Bhattacharjee et~al.(2023)Bhattacharjee, Kumarage, Moraffah, and Liu}]{Bhattacharjee2023}
Amrita Bhattacharjee, Tharindu Kumarage, Raha Moraffah, and Huan Liu. 2023.
\newblock \href {https://doi.org/10.18653/v1/2023.ijcnlp-main.40} {{C}on{DA}: Contrastive domain adaptation for {AI}-generated text detection}.
\newblock In \emph{Proceedings of the 13th International Joint Conference on Natural Language Processing and the 3rd Conference of the Asia-Pacific Chapter of the Association for Computational Linguistics (Volume 1: Long Papers)}, pages 598--610, Nusa Dua, Bali. Association for Computational Linguistics.

\bibitem[{Chen et~al.(2024)Chen, Qadri, Wen, Jain, Kirchenbauer, Zhou, and Goldstein}]{chen2024genqageneratingmillionsinstructions}
Jiuhai Chen, Rifaa Qadri, Yuxin Wen, Neel Jain, John Kirchenbauer, Tianyi Zhou, and Tom Goldstein. 2024.
\newblock \href {https://arxiv.org/abs/2406.10323} {Genqa: Generating millions of instructions from a handful of prompts}.
\newblock \emph{Preprint}, arXiv:2406.10323.

\bibitem[{Darda et~al.(2023)Darda, Carre, and Cross}]{Darda2023}
Kohinoor~Monish Darda, Marion Carre, and Emily~S. Cross. 2023.
\newblock \href {https://doi.org/10.1098/rsos.220915} {Value attributed to text-based archives generated by artificial intelligence}.
\newblock \emph{Royal Society Open Science}, 10(2):220915.

\bibitem[{Dugan et~al.(2024)Dugan, Hwang, Trhl{\'\i}k, Zhu, Ludan, Xu, Ippolito, and Callison-Burch}]{dugan-etal-2024-raid}
Liam Dugan, Alyssa Hwang, Filip Trhl{\'\i}k, Andrew Zhu, Josh~Magnus Ludan, Hainiu Xu, Daphne Ippolito, and Chris Callison-Burch. 2024.
\newblock \href {https://doi.org/10.18653/v1/2024.acl-long.674} {{RAID}: A shared benchmark for robust evaluation of machine-generated text detectors}.
\newblock In \emph{Proceedings of the 62nd Annual Meeting of the Association for Computational Linguistics (Volume 1: Long Papers)}, pages 12463--12492, Bangkok, Thailand. Association for Computational Linguistics.

\bibitem[{Gall{\'e} et~al.(2021)Gall{\'e}, Rozen, Kruszewski, and Elsahar}]{Galle2021}
Matthias Gall{\'e}, Jos Rozen, Germ{\'a}n Kruszewski, and Hady Elsahar. 2021.
\newblock \href {https://arxiv.org/abs/2111.02878} {Unsupervised and distributional detection of machine-generated text}.
\newblock \emph{arXiv preprint arXiv:2111.02878}.

\bibitem[{Gharib and Elgendy(2025)}]{genai-detect:2025:task:RA}
Rana Gharib and Ahmed Elgendy. 2025.
\newblock {RA at GenAI Detection Task 2: Fine-tuned Language Models For Detection of Academic Authenticity, Results and Thoughts}.
\newblock In \emph{Proceedings of the 1st Workshop on GenAI Content Detection (GenAIDetect)}, Abu Dhabi, UAE. International Conference on Computational Linguistics.

\bibitem[{Guo et~al.(2023)Guo, Zhang, Wang, Jiang, Nie, Ding, Yue, and Wu}]{guo2023close}
Biyang Guo, Xin Zhang, Ziyuan Wang, Minqi Jiang, Jinran Nie, Yuxuan Ding, Jianwei Yue, and Yupeng Wu. 2023.
\newblock How close is chatgpt to human experts? comparison corpus, evaluation, and detection.
\newblock \emph{arXiv preprint arXiv:2301.07597}.

\bibitem[{Hanley and Durumeric(2024)}]{hanley2024machine}
Hans~WA Hanley and Zakir Durumeric. 2024.
\newblock Machine-made media: Monitoring the mobilization of machine-generated articles on misinformation and mainstream news websites.
\newblock In \emph{Proceedings of the International AAAI Conference on Web and Social Media}, volume~18, pages 542--556.

\bibitem[{He et~al.(2022)He, Xu, Lyu, Wu, and Wang}]{He_Xu_Lyu_Wu_Wang_2022}
Xuanli He, Qiongkai Xu, Lingjuan Lyu, Fangzhao Wu, and Chenguang Wang. 2022.
\newblock \href {https://doi.org/10.1609/aaai.v36i10.21321} {Protecting intellectual property of language generation apis with lexical watermark}.
\newblock \emph{Proceedings of the AAAI Conference on Artificial Intelligence}, 36(10):10758--10766.

\bibitem[{Indurthi and Varma(2025)}]{genai-detect:2025:task2:Tesla}
Vijayasaradhi Indurthi and Vasudeva Varma. 2025.
\newblock {Tesla at GenAI Content Detection Task 1: LLM Agents in Multilingual Machine-Generated Text Detection}.
\newblock In \emph{Proceedings of the 1st Workshop on GenAI Content Detection (GenAIDetect)}, Abu Dhabi, UAE. International Conference on Computational Linguistics.

\bibitem[{Kaijie et~al.(2025)Kaijie, Xingyu, Shixuan, Sifan, Zikang, Benfeng, Licheng, Quan, Yongdong, and Zhendong}]{genai-detect:2025:task:CMI-AIGCX}
Jiao Kaijie, Yao Xingyu, Ma~Shixuan, Fang Sifan, Guo Zikang, Xu~Benfeng, Zhang Licheng, Wang Quan, Zhang Yongdong, and Mao Zhendong. 2025.
\newblock {CMI-AIGCX at GenAI Detection Task 2: Leveraging Multilingual Proxy LLMs for Machine-Generated Text Detection in Academic Essays}.
\newblock In \emph{Proceedings of the 1st Workshop on GenAI Content Detection (GenAIDetect)}, Abu Dhabi, UAE. International Conference on Computational Linguistics.

\bibitem[{Kirchenbauer et~al.(2023)Kirchenbauer, Geiping, Wen, Katz, Miers, and Goldstein}]{kirchenbauer2023watermark}
John Kirchenbauer, Jonas Geiping, Yuxin Wen, Jonathan Katz, Ian Miers, and Tom Goldstein. 2023.
\newblock A watermark for large language models.
\newblock In \emph{International Conference on Machine Learning}, pages 17061--17084. PMLR.

\bibitem[{Liang et~al.(2023)Liang, Yuksekgonul, Mao, Wu, and Zou}]{Liang2023}
Weixin Liang, Mert Yuksekgonul, Yining Mao, Eric Wu, and James Zou. 2023.
\newblock {GPT} detectors are biased against non-native english writers.
\newblock \emph{Patterns}, 4(7):100779.

\bibitem[{Mikros et~al.(2023)Mikros, Koursaris, Bilianos, and Markopoulos}]{Mikros2023}
George Mikros, Athanasios Koursaris, Dimitrios Bilianos, and George Markopoulos. 2023.
\newblock \href {https://ceur-ws.org/Vol-3496/autextification-paper9.pdf} {{AI}-writing detection using an ensemble of transformers and stylometric features}.
\newblock In \emph{Proceedings of the Iberian Languages Evaluation Forum ({IberLEF} 2023) co-located with the Conference of the Spanish Society for Natural Language Processing ({SEPLN} 2023)}, volume 3496 of \emph{CEUR Workshop Proceedings}, pages 1--14, Ja{\'e}n, Spain.

\bibitem[{OpenAI(2024)}]{openai2024gpt4omini}
OpenAI. 2024.
\newblock \href {https://openai.com/index/gpt-4o-mini-advancing-cost-efficient-intelligence/} {Gpt-4o mini: Advancing cost-efficient intelligence}.
\newblock \emph{OpenAI Blog}.

\bibitem[{Perkins et~al.(2024)Perkins, Roe, Vu, Postma, Hickerson, McGaughran, and Khuat}]{Perkins2024}
Mike Perkins, Jasper Roe, Binh~H. Vu, Darius Postma, Don Hickerson, James McGaughran, and Huy~Q. Khuat. 2024.
\newblock \href {https://doi.org/10.1186/s41239-024-00487-w} {Simple techniques to bypass {GenAI} text detectors: Implications for inclusive education}.
\newblock \emph{International Journal of Educational Technology in Higher Education}, 21(1):53.

\bibitem[{Rashidi et~al.(2023)Rashidi, Fennell, Albahra, Hu, and Gorbett}]{Rashidi2023}
Hooman~H. Rashidi, Brandon~D. Fennell, Samer Albahra, Bo~Hu, and Tom Gorbett. 2023.
\newblock \href {https://doi.org/10.1016/j.jpi.2023.100342} {The chatgpt conundrum: Human-generated scientific manuscripts misidentified as ai creations by ai text detection tool}.
\newblock \emph{Journal of Pathology Informatics}, 14:100342.

\bibitem[{Shah et~al.(2023)Shah, Ranka, Dedhia, Prasad, Muni, and Bhowmick}]{Shah2023}
Aditya Shah, Prateek Ranka, Urmi Dedhia, Shruti Prasad, Siddhi Muni, and Kiran Bhowmick. 2023.
\newblock \href {https://doi.org/10.14569/ijacsa.2023.01410110} {Detecting and unmasking {AI}-generated texts through explainable artificial intelligence using stylistic features}.
\newblock \emph{International Journal of Advanced Computer Science and Applications}, 14(10):1--10.

\bibitem[{Szyller et~al.(2021)Szyller, Atli, Marchal, and Asokan}]{10.1145/3474085.3475591}
Sebastian Szyller, Buse~Gul Atli, Samuel Marchal, and N.~Asokan. 2021.
\newblock \href {https://doi.org/10.1145/3474085.3475591} {Dawn: Dynamic adversarial watermarking of neural networks}.
\newblock In \emph{Proceedings of the 29th ACM International Conference on Multimedia}, MM '21, page 4417–4425, New York, NY, USA. Association for Computing Machinery.

\bibitem[{Tang et~al.(2024)Tang, Chuang, and Hu}]{tang2024science}
Ruixiang Tang, Yu-Neng Chuang, and Xia Hu. 2024.
\newblock The science of detecting llm-generated text.
\newblock \emph{Communications of the ACM}, 67(4):50--59.

\bibitem[{Team(2024)}]{gemini2024}
Gemini Team. 2024.
\newblock \href {https://arxiv.org/abs/2403.05530} {Gemini 1.5: Unlocking multimodal understanding across millions of tokens of context}.
\newblock \emph{arXiv preprint arXiv:2403.05530}.

\bibitem[{Vora et~al.(2023)Vora, Savla, Mehta, and Gawade}]{Vora2023}
Vismay Vora, Jenil Savla, Deevya Mehta, and Aruna Gawade. 2023.
\newblock \href {https://doi.org/10.17762/ijritcc.v11i9.8861} {A multimodal approach for detecting {AI} generated content using {BERT} and {CNN}}.
\newblock \emph{International Journal on Recent and Innovation Trends in Computing and Communication}, 11(9):691--701.

\bibitem[{Wang et~al.(2024{\natexlab{a}})Wang, Mansurov, Ivanov, Su, Shelmanov, Tsvigun, Afzal, Mahmoud, Puccetti, Arnold, Arnold, Whitehouse, Aji, Habash, Gurevych, and Nakov}]{wang2024semeval}
Yuxia Wang, Jonibek Mansurov, Petar Ivanov, Jinyan Su, Artem Shelmanov, Akim Tsvigun, Osama~Mohammed Afzal, Tarek Mahmoud, Giovanni Puccetti, Thomas Arnold, Thomas Arnold, Chenxi Whitehouse, Alham~Fikri Aji, Nizar Habash, Iryna Gurevych, and Preslav Nakov. 2024{\natexlab{a}}.
\newblock \href {https://doi.org/10.18653/v1/2024.semeval-1.279} {{S}em{E}val-2024 {Task 8}: Multidomain, multimodel and multilingual machine-generated text detection}.
\newblock In \emph{Proceedings of the 18th International Workshop on Semantic Evaluation (SemEval-2024)}, pages 2057--2079, Mexico City, Mexico. Association for Computational Linguistics.

\bibitem[{Wang et~al.(2024{\natexlab{b}})Wang, Mansurov, Ivanov, Su, Shelmanov, Tsvigun, Whitehouse, Mohammed~Afzal, Mahmoud, Sasaki, Arnold, Aji, Habash, Gurevych, and Nakov}]{wang-etal-2024-m4}
Yuxia Wang, Jonibek Mansurov, Petar Ivanov, Jinyan Su, Artem Shelmanov, Akim Tsvigun, Chenxi Whitehouse, Osama Mohammed~Afzal, Tarek Mahmoud, Toru Sasaki, Thomas Arnold, Alham Aji, Nizar Habash, Iryna Gurevych, and Preslav Nakov. 2024{\natexlab{b}}.
\newblock \href {https://aclanthology.org/2024.eacl-long.83} {M4: Multi-generator, multi-domain, and multi-lingual black-box machine-generated text detection}.
\newblock In \emph{Proceedings of the 18th Conference of the European Chapter of the Association for Computational Linguistics (Volume 1: Long Papers)}, pages 1369--1407, St. Julian{'}s, Malta. Association for Computational Linguistics.

\bibitem[{Weber-Wulff et~al.(2023)Weber-Wulff, Anohina-Naumeca, Bjelobaba, Folt\'{y}nek, Guerrero-Dib, Popoola, \v{S}igut, and Waddington}]{WeberWulff2023}
Debora Weber-Wulff, Alla Anohina-Naumeca, Sonja Bjelobaba, Tom\'{a}\v{s} Folt\'{y}nek, Jean Guerrero-Dib, Olumide Popoola, Petr \v{S}igut, and Lorna Waddington. 2023.
\newblock \href {https://doi.org/10.1007/s40979-023-00146-z} {Testing of detection tools for {AI}-generated text}.
\newblock \emph{International Journal for Educational Integrity}, 19(1):1--39.

\bibitem[{Wu and Flanagan(2023)}]{Wu2023}
Hongyu Wu and Tom Flanagan. 2023.
\newblock \href {https://doi.org/10.47611/jsrhs.v12i3.5064} {The limits of {AI} content detectors}.
\newblock \emph{Journal of Student Research}, 12(3):1--7.

\bibitem[{Wu et~al.(2023)Wu, Yang, Zhan, Yuan, Wong, and Chao}]{wu2023survey}
Junchao Wu, Shu Yang, Runzhe Zhan, Yulin Yuan, Derek~F Wong, and Lidia~S Chao. 2023.
\newblock A survey on llm-gernerated text detection: Necessity, methods, and future directions.
\newblock \emph{arXiv preprint arXiv:2310.14724}.

\bibitem[{Yu et~al.(2023)Yu, Chen, Feng, and Xia}]{yu2023cheat}
Peipeng Yu, Jiahan Chen, Xuan Feng, and Zhihua Xia. 2023.
\newblock Cheat: A large-scale dataset for detecting chatgpt-written abstracts.
\newblock \emph{arXiv preprint arXiv:2304.12008}.

\bibitem[{Zaghouani et~al.(2024)Zaghouani, Ahmed, Zhang, and Rezk}]{zaghouani-etal-2024-qcaw}
Wajdi Zaghouani, Abdelhamid Ahmed, Xiao Zhang, and Lameya Rezk. 2024.
\newblock \href {https://aclanthology.org/2024.lrec-main.1172} {{QCAW} 1.0: Building a qatari corpus of student argumentative writing}.
\newblock In \emph{Proceedings of the 2024 Joint International Conference on Computational Linguistics, Language Resources and Evaluation (LREC-COLING 2024)}, pages 13382--13394, Torino, Italia. ELRA and ICCL.

\bibitem[{Zaitsu and Jin(2023)}]{Zaitsu2023}
Wataru Zaitsu and Mingzhe Jin. 2023.
\newblock \href {https://doi.org/10.1371/journal.pone.0288453} {Distinguishing chatgpt(-3.5, -4)-generated and human-written papers through japanese stylometric analysis}.
\newblock \emph{PLOS ONE}, 18(8):1--12.

\end{thebibliography}

\appendix



\end{document}